# The Strain of Success: A Predictive Model for Injury Risk Mitigation and Team Success in Soccer

Gregory Everett[1], Ryan Beal[2], Tim Matthews[2], Timothy J. Norman[1], Sarvapali D. Ramchurn[1]

[1]University of Southampton, Southampton, UK
[2]Sentient Sports, UK

## 1. Introduction

The prevalence and impact of player injuries in soccer are profound and multifaceted, influencing team performance, financial stability, and player welfare. In the 2021/22 season, the 'Big Five' European soccer leagues experienced a staggering £513 million in injury-related costs, a 29% increase from the previous season, as reported by Howden, an insurance brokerage firm.[1] Research indicates that injury rates during soccer matches can vary from 10 to 60 incidents per 1000 playing hours [1], with a marked impact on team outcomes in both league play and cup competitions. These findings underscore the necessity of prioritizing player welfare, not only as a matter of ethical responsibility but also as a critical element in a team's strategic planning and overall success.

Soccer teams are increasingly leveraging sports science and AI, particularly in using game and training data to predict injuries. However, the application of AI for long-term player welfare strategies remains underexplored. Current team management often follows a short-term, "next game" focus, prioritizing immediate performance over long-term player health. This approach, while addressing immediate pressures for success, overlooks the risks of player fatigue and injury, highlighting the need for more holistic, future-oriented team management strategies.

In this paper, we explore the balance of risk and reward in team selections during congested soccer season schedules. Drawing inspiration from multi-agent systems in team formation literature, we address the challenge of selecting teams from a dynamic pool of players to maximize performance while minimizing injury risks. We propose a novel sequential team selection model that utilizes real-world data to assess injury probabilities and impacts. This model, framed as a Markov decision process (MDP) and optimized using Monte-Carlo tree search (MCTS), aims to balance team performance with the risk of long-term player unavailability due to injuries. We analyze this model using data from 760 English Premier League games across two seasons and compare its performance against a myopic Greedy strategy.[2] Our approach not only offers insights for soccer but also has potential applications in other team sports with high injury risks.

Therefore, we have developed a data-driven team selection strategy that considers player injury risk and match outcome probabilities, reducing player injuries while sustaining long-term team success. In summary, we make the following novel contributions:

- A probabilistic player injury prediction model that outperforms numerous baselines.
- A forward-looking team formation model which aims to maximize long-term reward (points gained) while minimizing player injury risk. This recommends when to rest fatigued players at the optimal times.

---

[1] [Europe's top clubs paid record-high prices for injuries in 2021-22, study reveals - The Athletic](#)
[2] Open-source code is available at: https://github.com/Sentient-Sports/Strain-of-Success

- Our model achieves similar performance to greedy team selections whilst reducing the incidence of player injury by ~13% and the standard deviation in season-expected points by ~17%.
- We use data to demonstrate that managers most commonly follow a greedy approach during team selection.

In Section 2, we discuss related work. Section 3 presents an overview of the team selection model. Section 4 explains the injury probability model and Section 5 presents the match outcome prediction model. In Section 6, we formulate team selection as an MDP and select optimal teams using MCTS. Section 7 evaluates our approach and findings are discussed in Section 8. Finally, Section 9 concludes.

## 2. Related Work

Injuries in sports, particularly their frequency in matches and their impact on team performance, have been extensively studied [2]. Research has also focused on the factors influencing injury risk, including workload [3], previous injuries [4], and weather conditions [5]. However, there's a gap in research regarding optimizing team selections to mitigate long-term injury impacts on performance and welfare.

Team selection optimization has been explored in some studies. Beal et al. [6] used network analysis to understand player interactions and select teams based on teamwork metrics, whereas Matthews et al. [7] employed Bayesian reinforcement learning for selecting optimal fantasy soccer teams. These studies, while effective in evaluating player value, overlook the impact of fatigue and injury on long-term performance.

Our approach also considers the lack of models in multi-agent system literature that account for agent fault (injury) probability and the risk-reward tradeoff in team formation. Existing models focus on dynamic task environments or agents selecting teams based on personal value functions [8, 9]. Some studies have addressed agent faults in team formation, like robust team formation [10] and recoverable team formation [11], or frameworks for simulated scenarios where team members are replaced upon failure [12]. In contrast, our approach proactively reduces fault probability.

We believe our work is the first to investigate long-term team selection considering the extended risk and reward of player availability and performance. The following section outlines our team selection framework.

## 3. A Model to Optimise On-Pitch Success While Minimising Injury Risk

In this section, we present an overview of the team selection optimization framework and provide mathematical definitions for the foundations of the team selection problem. The problem of team selection necessitates consideration of many factors such as the combined ability of the team, the importance of the upcoming match and the risk and impact of long-term player injuries if they play. We provide an overview of the team selection model in the next subsection.

## 3.1. Team Selection Model Overview

Here, we provide an overview of the team selection framework used throughout this paper. Figure 1 illustrates the process of team selection over a season, which we describe step by step below.

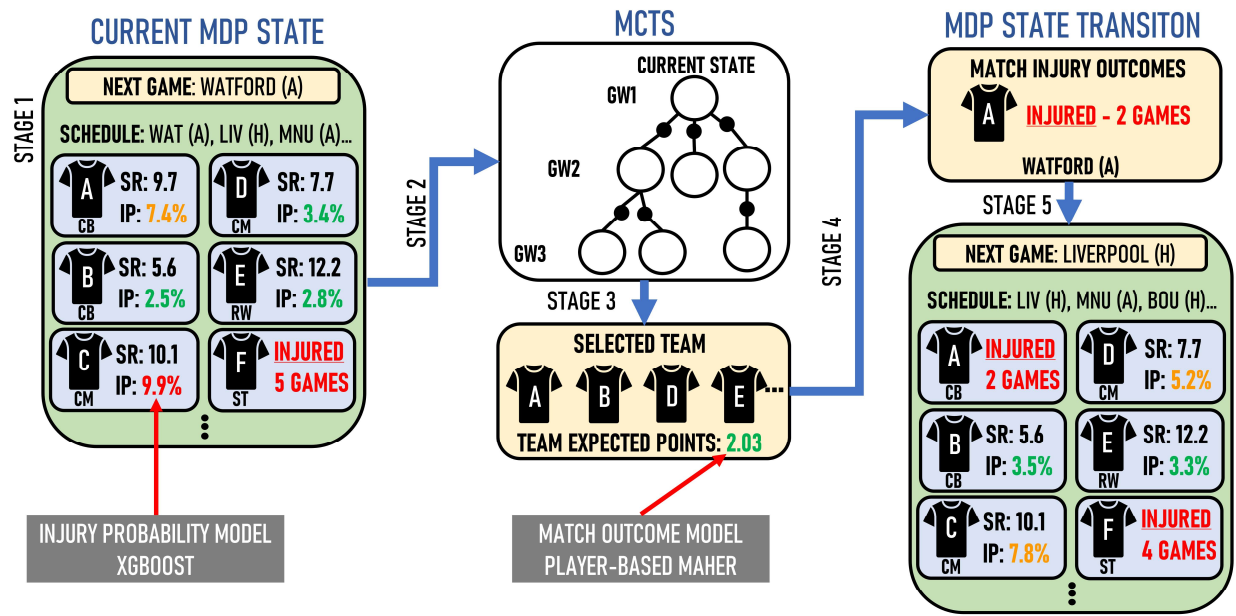

**Figure 1** – Team selection framework example showing how squad updates occur after a game in our model (where shirts represent players). SR=Skill Rating, IP=Injury Probability.

1. **Current Team State –** Before a game, our model evaluates the squad, considering each player's skill rating based on past performance (see Section 5) and their injury probability for the next game, as determined by an injury probability model (see Section 4). Additionally, information about the upcoming opponent and future fixtures is taken into account. This collective data forms the Markov decision process (MDP) state.
2. **Manager Decision-Making –** The goal of the optimiser is to select the optimal team for maximizing long-term performance. Using Monte Carlo tree search, the model approximates the long-term values of different team compositions, balancing immediate rewards and future injury risks. This process is aided by a player-based match prediction model, which estimates the potential points from a game.
3. **Team Selection –** Based on the MCTS analysis, the model selects the optimal team, aiming to achieve the predicted immediate rewards from the game.
4. **MDP State Transition –** Following the game, player injury probabilities are updated, and any new injuries are recorded, reflecting changes in the team's condition.
5. **New MDP State –** The new MDP state incorporates these updates, including player injuries and revised probabilities, as well as information about the next upcoming game.

## 3.2. Defining the Mathematical Model

In this subsection, we mathematically model players and team selection, and then introduce player availability and injury.

### 3.2.1. Players and Games

We have a set of $N$ players defined as $A = \{a_1, a_2 \ldots, a_N\}$. A sequential set (season) of $K$ games, is also defined as $T = \{t_1, \ldots, t_K\}$. For this problem, we study the English Premier League, so the length of the season is $K = 38$ games. The manager must choose a team of players from $A$ to play each game $t \in T$. We define a selected team as $C$ where $C \subseteq A$, and for a soccer game, $C$ must contain 11 players. Each player $a \in A$ has a skill value that quantifies their impact towards winning soccer games. We define the skill value of a player $a$ as $\Omega_a \in \boldsymbol{\Omega}$ where $\boldsymbol{\Omega}$ is the set of all player skill values. These skill values are learned from past games that involved that player.

Each game $t \in T$ has a difficulty value defined as $\alpha_t \in \boldsymbol{\alpha}$, representing the difficulty of the game based on the opponent and venue (i.e., home or away). As this difficulty refers to the strength of the opponent, it is a constant value independent of the selected team $C$. Given $\alpha_t$ and a team $C$, we define the value of the team $C$ for game $t$ as $V(C, t) = f(C, \alpha_t)$ where $f$ is a model that computes the team's value as the expected points received for the team in that game. The model $f$ and game difficulties $\boldsymbol{\alpha}$ are trained and predicted using past game data (expanded on in Section 5).

### 3.2.2. Player Availability

Every player has an injury probability at each game. We define a matrix of injury probabilities $\Theta$ where $\Theta_{a,t}$ is the injury probability of player $a$ at game $t$. After each game, injury probabilities are updated using a statistical model described in Section 4. The set of possible injury outcomes of a game $t$ is defined as $o_t \in O_t$, where $o_t$ contains the occurrences of new injuries and their lengths for every player resulting from the game $t$. The probability of an outcome at a game $t$ depends on the players selected to play (i.e., $C$) and their injury probabilities, $\Pr(o_t | C, \Theta_t)$.

Every game $t \in T$ has a timestep $\tau_t$ starting from the initial game which has a timestep 0. These timesteps represent days. For example, assuming the first game of the season has timestep $\tau_0 = 0$, and the next game is four days later, the timestep for the next game would be $\tau_1 = 4$. An unavailability length vector is defined for each game $t \in T$ as $L_t = \{l_1, l_2, \ldots, l_N\}$ where $l_a$ is the number of games that the player $a$ is unavailable for through injury (starting from game $t$). Naturally, $l_a$ is zero when a player is not injured. A player $a$ cannot be selected in the team $C$ if $l_a > 0$ (i.e., $\forall a \in C, l_a = 0$). Unavailability length reduces by one after each game for an injured player until it reaches zero and the player becomes available to play again. The unavailability length vector updates after each game $t$ for players in $C$ given the outcome of the game $o_t$. Injury duration is learned from a distribution of injury lengths in past data and is converted to a game count based on the number of games whose timesteps fall within the injury period (see Section 4). In the next subsection, we expand on our player injury prediction model.

## 4. Calculating Player Injury Risk

In this section, we present an injury risk model for soccer and explore the most contributory features towards increased player injury risk.

### 4.1. Injury Risk Model

The initial step for our model is the selection of features that appropriately summarize the factors that influence player injury risk. The diagram of the injury model and the features it is trained on are shown in Figure 2.

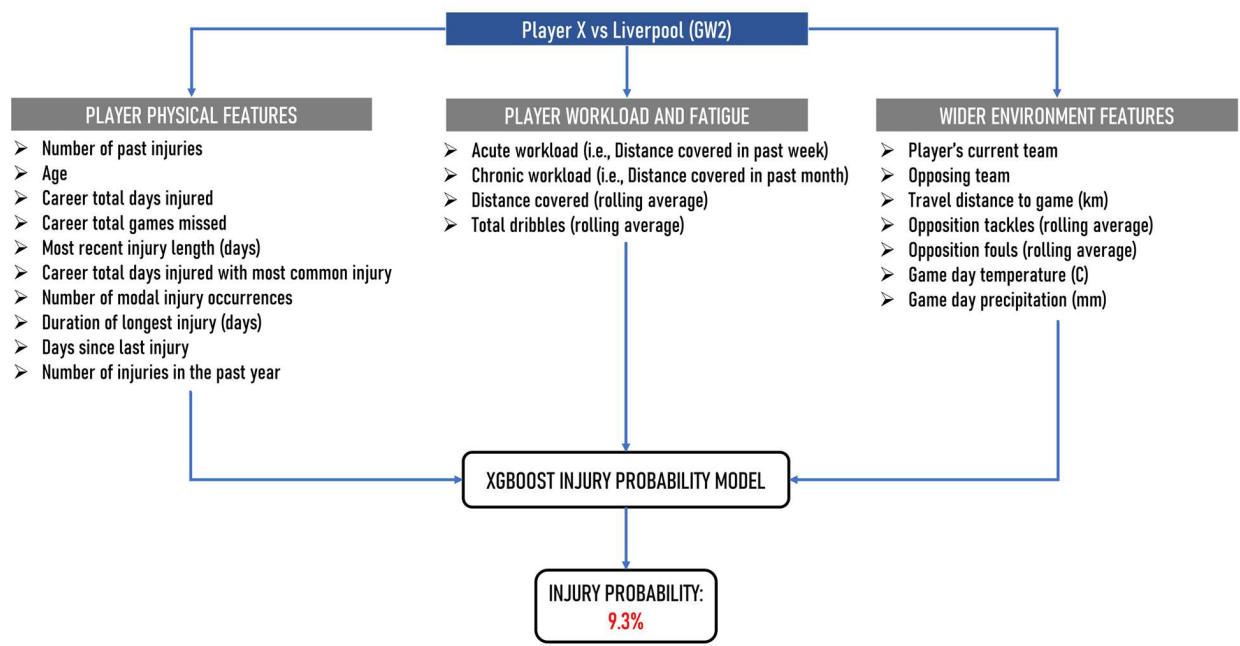

Figure 2 - Player injury probability model features and framework.

Prior work has identified relationships between these features and injury risk for athletes in sports [3, 5, 4]. At each timestep $t$, we use our feature set, denoted as risk factors $F_t$, as input to a model to predict injury probabilities, $\Theta_t$. We use XGBoost due to its strong performance on tabular data and train the model with a binary cross-entropy loss function so we can obtain well-calibrated probabilities. Ground truth data to train this model is described in Section 7. Computing physical metrics that are used to calculate fatigue-related features typically requires expensive tracking systems. To make our team formation model more accessible, we estimate player physical metrics using cheaper on-ball event data for past games from that player in the 2017/18 and 2018/19 English Premier League seasons, and a player location prediction model [13] that predicts tracking data from event data using a combination of graph neural networks and long-short term memory components. When we compute player workload and fatigue features in $F_t$ during our simulation of a season, we extrapolate the mean distance covered and dribbles per game for individual players from past game data to estimate these features. Player physical features are also updated during simulation.

With our team selection model, we can control the workload and fatigue features of players based on the selection of that player for upcoming games. For a single player, the goal is to maximise their output to the team whilst limiting the risk of them fatiguing and getting injured. Figure 3 provides an illustrative example of balancing player workload and injury risk over a run of games.

Once we have used the model above to sample player injury, we then sample the length of that injury from a Gaussian distribution fit to the total population of injury lengths from our dataset (described in Section 7). When an injury occurs, $L_t$ is updated with a sample from the Gaussian distribution denoting the number of games within the injury period. Any negative samples from the Gaussian distribution are assumed to be an injury length of 0 games.

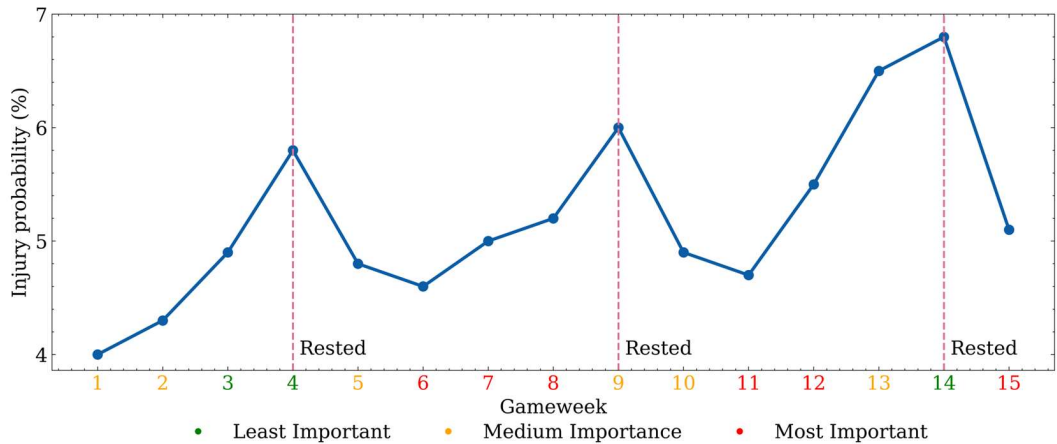

Figure 3 - Example diagram of the injury probability of a single player over time and the process of balancing player impact with the regenerative effects of resting the player. Color-coded gameweeks represent match importance. Red lines determine games where the player is rested – the player is selected for all other games.

## 4.2. **Feature Contributions to Injury Risk**

We assess each feature's contribution to our model using SHAP values [14]. Figure 10 in the Appendix lists the full analysis of feature contribution towards injury risk in players for our dataset. Features such as team form, days since the previous game, and other physical features were also curated. However, these features had minimal performance impact, with some already incorporated in other interrelated features. As a result, we removed them from the final feature set. We find that the most contributory feature was acute workload with a mean SHAP value of 0.0033, followed by the number of past injuries (0.0029), career total days injured (0.0023) and distance covered (0.0022). This shows that a combination of both workload and player injury history is important towards a player's injury risk.

We also explore the impact of features on individual predictions. Figure 4 shows how a model prediction for a player's injury probability in an upcoming game can be decomposed to see the individual impact of certain features.

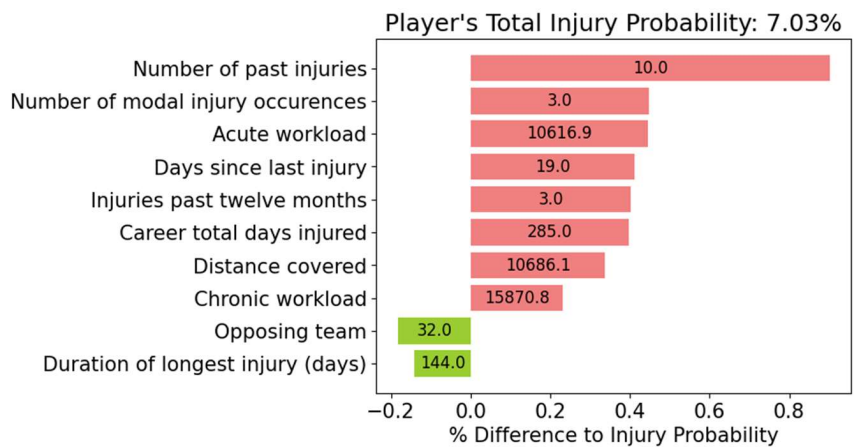

Figure 4 - Explainable Injury Predictions. For a single player before a game, their injury probability is computed and the most contributory features towards this are listed with their absolute impact on the injury probability and their value (in the bar).

This allows players and coaches to be better informed on why the model has flagged a high injury risk or suggested a player to be rested. Our model is designed to use key features from event and injury data, offering flexibility for customization. Clubs with access to more detailed health and training data on their players can adapt the feature input to enhance prediction accuracy and gain deeper insights into injury risks. This approach offers a versatile solution, particularly for clubs with limited data, providing a generalist framework that can be tailored to their specific needs.

## 5. Calculating Match Reward

In our team formation framework, the reward for a team in a single game $t$ is defined as $V(C, t) = f(C, \alpha_t)$. In soccer, $C$ is the selected players and $\alpha_t$ is the opponent's strength. We use these factors to predict match outcomes with a match prediction model $f$. Most previous match prediction models utilize historical club results, overlooking player-level contribution. We therefore adapt the well-established Maher prediction model [15], with proven success in betting applications, to predict match outcomes using player skill. The Maher model uses Poisson regression to predict the goals scored by each club in a game and compute match outcome probabilities, with club identities and a home parameter as features and coefficients learned from past games.

We modify the Maher model by replacing club identity features from the original paper with numeric player-based features. Specifically, we use players' total attacking and defending contributions in past games, quantified by the Valuing Actions by Estimating Probabilities (VAEP) [16] metric. The total VAEP (from past games) of a player is defined as their skill value $\Omega_a$. The summed skill values of all players in selected teams $i$ and $j$ yield team values $V_i$ and $V_j$, which are used as model features. The model learns the relationship between these features and team goals in past games. When predicting the match outcome for a game $t$, the opponent team strength, also defined as $\alpha_t$, is the opposition club's average team value $V_j$ across all past real-world games. The manager's team value is the total VAEP of the players selected in $C$. The model uses these features to compute goal probabilities for each team, and the win/draw/loss probability for each team is derived from the probability of each possible scoreline. Against this background, we define $V(C, t) = 3 \cdot \Pr(win|C) + \Pr(draw|C)$ where the reward is the expected points for the team using match outcome probabilities from the model $f$ given team $C$ plays.

When comparing the predictive performance of the player-level model to the original team-level Maher model using a 70/30 train/test split and five-fold cross-validation with random shuffling over our two-season dataset, we use log-loss to measure the calibration of match outcome probabilities to real match outcomes. It is found that the player-level model has a log-loss score of $0.915 \pm 0.020$ and the team-level model has a log-loss score of $0.910 \pm 0.018$. This shows that our player-level model achieves similar performance within error boundaries to a well-established prediction model, whilst providing additional flexibility in measuring outcome probabilities dependent on selected players.

## 6. Optimising Team Selections for Long-Term Performance

In this section, we frame our long-term team selection problem as a Markov decision process and explain how we utilize Monte Carlo Tree Search to optimise team selections to improve team performance with reduced player injuries.

## 6.1. Team Selection - A Markov Decision Process

We formulate our soccer team selection model as an MDP defined as $G = \langle S, \mathcal{A}, P, R, \gamma \rangle$ with a set of states $S$ and set of actions $\mathcal{A}$. For $s, s' \in S$ and $a \in \mathcal{A}$ we have a transition function $P = \Pr(s'|s,a)$ and a reward function $R(s,a)$. Finally, we have a discount factor $\gamma$. It is assumed that the manager has access to data on game difficulty, player injury risk factors and team values (i.e., player skills). Each MDP state contains:

- The current game $t$
- The set of remaining season games $Z_t$
- The set of injury risk factors $F_t$
- The set of current player injury probabilities $\Theta_t$
- The unavailability length vector $L_t$ (i.e., the team's injury list)

Where $Z_t \subseteq T$ and $Z_t$ holds all games after and including game $t$. The set of actions $\mathcal{A}$ is the set of all selectable teams such that $\forall a \in C, l_a = 0$ for all teams $C \in \mathcal{A}$. In other words, $\mathcal{A}$ is the set of all possible teams where all players in the team must be available to play (i.e., not injured). Clearly, the action set updates given the availability of players encapsulated in the current state. The reward for a game $t$ and team $C$, encapsulated in a state-action pair, is $V(C, t)$. As defined in Section 5, this reward is the expected points in the upcoming game.

At each state transition, $t$ and $Z_t$ update to the next game and the new set of remaining season games respectively with a probability of 1. The set of player risk factors, $F_t$, update given the chosen team $C$ with the updated workload for selected players, rest time given to non-selected players and new injuries. Furthermore, the set of injury probabilities is updated using the new risk factors and the model in Section 4. The unavailability length vector updates dependent on the outcome $\Pr(o_t|C, \Theta_t)$. After each game where a player $a$ is unavailable, their unavailability length decreases by 1. All states where the current game $t$ is the final game of the season $t_K$ are terminal states. Therefore, a walk from the initial state (i.e., game 1) to a terminal state contains $K-1$ transitions.

The objective is to select optimal teams using knowledge of player injury risk, game difficulty and team value. For each game $t \in T$, the decision-maker receives the state $(t, Z_t, F_t, \Theta_t, L_t)$ and aims to maximize the value of the Bellman equation:

$$V(s) = \max_a \left( R(s,a) + \gamma \sum_{s'} \Pr(s'|s,a) V(s') \right)$$

This optimization problem can be viewed as the manager competing against an adversarial environment. Player injury and unavailability is stochastically determined by the environment based on the manager's selected teams. The manager must balance immediate reward for the upcoming game with the risk and consequences of skilled players missing future games. This bears similarity to Expectimax problems, where optimal actions must consider all probabilistic state transitions and their expected values.

Due to a large action set resulting from many possible teams, and outcomes resulting from an action, the search tree is vast. For our soccer team selection problem, with the additional constraints we will describe in Section 6.2, there are up to ~2000 possible actions at each timestep and up to $38^{11}$ possible new states resulting from an action, accounting for injuries of varying length. Given this complexity, computing an exact solution for the Bellman equation is infeasible. To address this challenge, we use MCTS to approximate the Q-values of actions.

## 6.2. Team Selection Constraints

In soccer, a manager selects players whose roles fit a chosen formation. Each player $a \in A$ has a role $X_a$ which can be one of [GK,DEF,MID,ATT]. We express a generalized single-game constraint optimization problem in the equation below. The latter four constraints are a simplification that vastly reduces the action space to only plausible formations.[3] For simplicity, we assume that managers always choose their preferred formation which is learned from past data.

$$\begin{aligned}
\text{maximize} & \sum_{a \in C} \Omega_a \\
\text{subject to } & |C| = 11 \\
& \forall a \in C, l_a = 0 \\
& \sum_{a \in C} (X_a == GK) = 1 \\
3 \leq & \sum_{a \in C} (X_a == DEF) \leq 5 \\
3 \leq & \sum_{a \in C} (X_a == MID) \leq 5 \\
1 \leq & \sum_{a \in C} (X_a == ATT) \leq 3
\end{aligned}$$

This extends to the long-term optimization problem in soccer where selected teams must satisfy these constraints to optimise long-term rewards [17]. Our action space is limited to teams that fit the constraints above.

## 6.3. Sampling Optimal Team Selections

Monte-Carlo tree search is an anytime search algorithm for identifying optimal actions using a combination of heuristics and simulation to incrementally build an asymmetric search tree. It has considerable success in domains that are represented as trees of sequential decisions, such as Go [18] and Coalition Structure Generation [19]. We apply MCTS to our team selection problem by iterating through each game $t \in T$ and approximating the optimal action by guiding the search towards promising actions. As our MDP is stochastic, the search tree comprises decision nodes (i.e., an action must be chosen) and chance nodes (i.e., an outcome is selected by the environment with some probability). Assuming we are searching for the optimal action for the first game $t_1$, we define the search tree constructed by our MCTS algorithm as follows:

- The root decision node refers to the initial state of the MDP (i.e., $(t_1, Z_{t_1}, F_{t_1}, \Theta_{t_1}, L_{t_1})$).
- Each decision node encapsulates an MDP state $s \in S$.
- Each tree branch refers to an action chosen (i.e., $C$) by the manager at the current node.
- Each chance node refers to an outcome chosen by the environment given the connected decision node and branch.
- Terminal decision nodes refer to terminal MDP states.

---

[3] All formations seen in the data fit this formulation.

The maximum tree depth is $K$, a node on the $k$'th level of the tree contains a state associated with a game $t_k$, and a path from the root to a terminal node runs through all games.

### 6.3.1. MCTS Algorithm

For every iteration of the MCTS algorithm, there are four key steps: selection, expansion, simulation and backpropagation. Further details on these steps are as follows:

1) **Selection -** Select the most promising action from the root using the popular UCB1 strategy [20] until an unexplored or terminal node is reached. For each node $N_s$, with state $s$, their child node encapsulates new state $s'$ with probability determined by an action $a$ and the MDP's outcome probabilities $\Pr(o_t|C, \Theta_t)$.
2) **Expansion -** Expand node $N_s$ by randomly selecting an unexplored action. For our domain, the action space is large (up to ~2000 teams), and many actions may be highly unfavorable. To avoid inefficient computational resource allocation, we use progressive widening.
3) **Simulation -** After reaching a leaf node $N_s$, the objective is to approximate its value $V(s)$. We do this using an MDP rollout until a terminal state is reached and compute the cumulative reward. Typically, random actions (teams) are chosen during rollout, however, this strategy can lead to noisy approximations of $V(s)$. Furthermore, an action's impact on future player injury risk reduces greatly after a number of games look ahead. For our rollouts, at each game, the team $C$ is chosen to maximise the expected points for that game. The first three rollout state transitions occur normally and then the remaining games are simulated under the assumption of no further injuries. The resulting cumulative reward, outputted as the node value, provides stable approximations of an action's impact on the short-term injury risk and the effect of long-term player injury.
4) **Backpropagation -** After approximating the value of the newly added child node $N_s$, we backpropagate the value up the search tree until the root node is reached.

To address a large action space, we use progressive widening [21, 22]. The set of selectable actions is initially artificially limited in size and progressively expanded over time. This strategy improves the allocation of resources when running in a limited time by prioritizing promising actions. This helps to approximate the balance between immediate rewards from selecting players with the long-term reward of resting them. Actions are ranked using this and the action set progressively widens.

## 7. Evaluation

To evaluate our approach, we use on-ball events data for the 2017/18 and 2018/19 English Premier League (EPL) seasons[4] and an extensive player injury dataset for these seasons and historic career injuries.[5] These datasets allow us to assess the real-world value of our models.

### 7.1. Experiment 1: Performance Comparison of Injury Risk Model

In this experiment, we compare the performance of our XGBoost injury model to other models (using the same features) and a heuristic baseline that uses the injury rate of players in past games as predictions. We use a 70/30 train/test split and five-fold cross-validation with random shuffling for our 2017/18 and 2018/19 EPL injury dataset to evaluate these injury models. Optimal hyperparameters are selected for each model using grid search cross-validation. The XGBoost model achieved a log loss of 0.1676 ± 0.0005, outperforming Logistic Regression, Random Forest and Neural Network models which all achieved poorer log loss scores. The heuristic baseline achieved a log loss score of 0.1700 ± 0.0002.

---

[4] Sample versions of StatsBomb event datasets available at: https://github.com/statsbomb.
[5] https://www.transfermarkt.co.uk/

These results highlight the additional predictive benefit of player-based features to learn injury risk beyond just the heuristic baseline. Despite the challenging low injury rate in soccer (~4% in our dataset), XGBoost consistently outperforms the other methods across all five folds. Figure 5 assesses the long-term performance of the model by comparing the number of predicted and actual injuries for each team per season over the 2017/18 and 2018/19 EPL data.

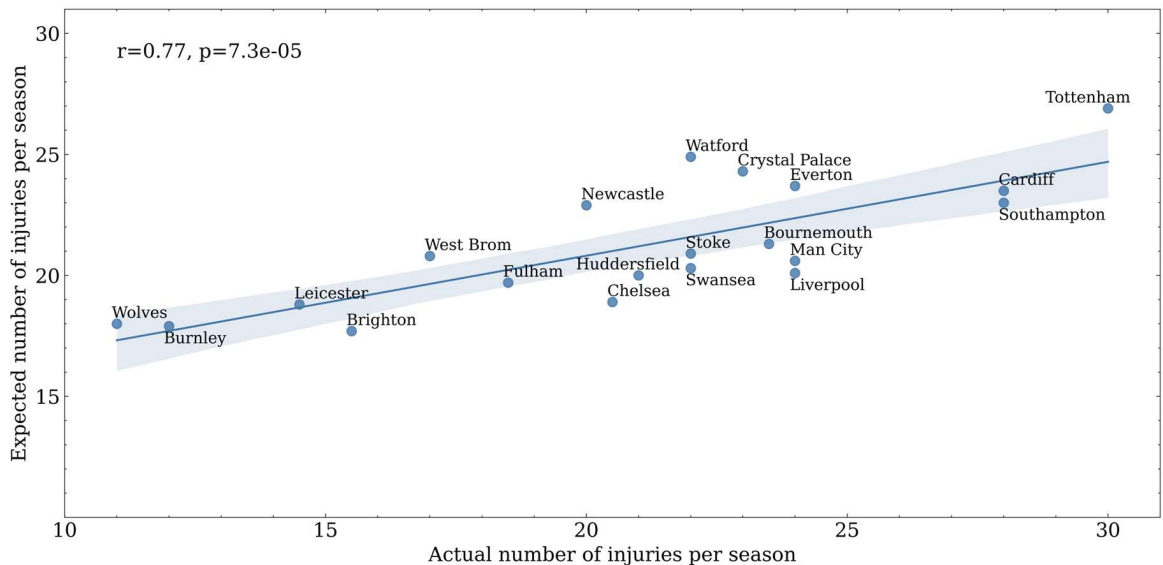

Figure 5 - Predicted number of injuries in a season (using the XGBoost injury model) for each team compared to actual number of injuries.

The mean percentage difference between the expected and actual number of injuries per team is ~14%. The Pearson correlation coefficient is 0.77, significant at $p<0.01$.

### 7.2. **Experiment 2: Model Comparison to Human Expert Selections**

Next, we compare our MCTS selections to real-world teams chosen by human-expert managers. Using real-world team states, including player injuries and workloads, we measure team similarity as the average percentage of shared players between real-world and MCTS-selected teams. Figure 6 summarizes the similarities for each strategy compared to manager selections across the league.

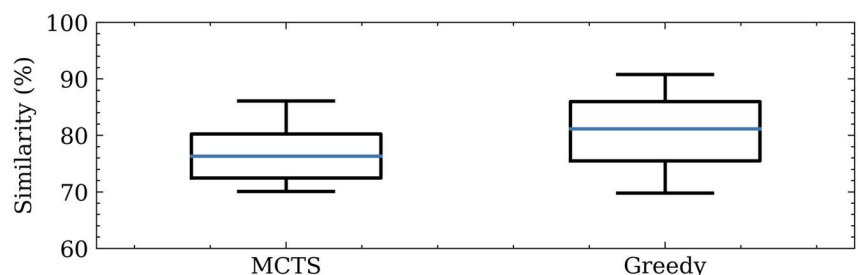

Figure 6 - Similarity comparisons between a manager's team selections and the MCTS and Greedy strategies.

Our analysis demonstrates that team selection processes by human expert managers in real-world settings closely mirror our greedy model, indicating a widespread managerial inclination toward prioritizing short-term results. This pattern is likely a consequence of the high-pressure environment faced by managers, where the emphasis on immediate success could inadvertently risk long-term player welfare. Conversely, our approach, utilizing the MCTS model, aims to strike a more balanced consideration between immediate performance and the sustained health and career longevity of players. Observing the widespread use of the greedy strategy in practice, we will adopt it as a comparative benchmark in our subsequent experiment. This will allow us to assess the practical efficacy of our data-driven approach against traditional managerial decision-making in real-world scenarios.

### 7.3. **Experiment 3: Model Season Performance**

We now evaluate the performance of an MCTS-based method against the greedy strategy commonly used in real-world scenarios which was highlighted in the previous experiment, designed to select teams anticipated to score the maximum points in the next game. We conduct separate complete season simulations using each strategy: 100 season simulations for each team in the 2018/19 EPL season using MCTS (due to computational cost - run on a compute cluster) and 1000 season simulations for each team using the greedy strategy. The outcomes of these simulations, comparing the two strategies across several metrics, are presented in Table 1.

Table 1 - Season Performance of MCTS compared to a Greedy strategy

| | Expected Points | | Std. Dev. | | Squad Injuries | | | Optimal Team Injuries | | |
|---|---|---|---|---|---|---|---|---|---|---|
| **Team** | **MCTS** | **Greedy** | **MCTS** | **Greedy** | **MCTS** | **Greedy** | **Dec (%)** | **MCTS** | **Greedy** | **Dec (%)** |
| Arsenal | 58.7±0.6 | 58.7±0.3 | 3.0 | 4.1 | 24.7 | 25.9 | 4.7 | 16.5 | 19.0 | 12.8 |
| Bournemouth | 54.4±0.5 | 54.3±0.2 | 2.8 | 3.7 | 19.8 | 20.1 | 1.6 | 13.6 | 15.3 | 11.5 |
| Brighton | 42.6±0.4 | 42.1±0.1 | 2.1 | 2.3 | 16.1 | 19.1 | 15.5 | 11.2 | 14.6 | 23.3 |
| Burnley | 47.7±0.4 | 48.0±0.2 | 2.3 | 2.8 | 17.8 | 19.1 | 6.7 | 13.0 | 14.9 | 12.9 |
| Cardiff | 32.5±0.3 | 32.4±0.1 | 1.3 | 1.6 | 18.9 | 20.0 | 5.2 | 13.6 | 15.5 | 12.0 |
| Chelsea | 70.0±0.7 | 69.5±0.3 | 3.8 | 5.0 | 21.8 | 22.0 | 0.5 | 14.9 | 16.1 | 7.5 |
| Crystal Palace | 55.7±0.8 | 55.5±0.3 | 4.2 | 4.3 | 19.4 | 20.2 | 4.3 | 12.9 | 15.3 | 15.5 |
| Everton | 49.2±0.5 | 49.4±0.1 | 2.4 | 3.1 | 21.1 | 21.9 | 3.3 | 14.5 | 16.2 | 11.6 |
| Fulham | 30.3±0.2 | 30.5±0.1 | 1.1 | 1.3 | 19.0 | 19.7 | 3.3 | 13.1 | 14.9 | 12.2 |
| Huddersfield | 39.5±0.4 | 39.5±0.1 | 1.9 | 2.1 | 18.9 | 20.1 | 5.8 | 13.2 | 15.7 | 15.8 |
| Leicester | 49.1±0.5 | 49.4±0.2 | 2.6 | 2.9 | 17.5 | 18.6 | 5.8 | 12.7 | 14.4 | 11.4 |
| Liverpool | 79.7±1.1 | 79.5±0.4 | 5.1 | 5.3 | 23.0 | 23.7 | 2.8 | 15.2 | 17.4 | 12.7 |
| Man City | 87.7±1.1 | 87.6±0.4 | 5.5 | 5.7 | 22.7 | 23.4 | 3.0 | 16.0 | 17.7 | 9.7 |
| Man United | 62.5±0.6 | 62.7±0.2 | 3.2 | 3.9 | 24.6 | 25.7 | 4.3 | 16.2 | 18.9 | 14.5 |
| Newcastle | 44.3±0.4 | 44.3±0.2 | 1.9 | 2.4 | 20.0 | 20.6 | 2.7 | 13.7 | 15.7 | 12.2 |
| Southampton | 44.6±0.3 | 44.5±0.1 | 1.8 | 2.0 | 18.8 | 19.2 | 1.8 | 12.5 | 14.5 | 13.8 |
| Tottenham | 69.1±0.9 | 68.8±0.2 | 4.5 | 4.8 | 21.7 | 22.6 | 3.6 | 14.8 | 16.8 | 11.7 |
| Watford | 56.5±0.6 | 57.1±0.3 | 3.5 | 3.8 | 18.9 | 20.4 | 5.8 | 13.4 | 15.5 | 13.5 |
| West Ham | 52.0±0.6 | 52.6±0.2 | 3.1 | 3.3 | 22.3 | 23.2 | 3.6 | 14.4 | 16.8 | 13.9 |
| Wolves | 35.0±0.3 | 35.3±0.1 | 1.7 | 1.9 | 16.1 | 17.3 | 6.7 | 11.9 | 13.6 | 13.0 |

Our MCTS approach closely matched the greedy strategy in expected seasonal points, with only a -0.1% ± 0.3 mean difference. It notably reduced squad injuries by about 5% and injuries to the optimal team, which is the top 11 players (in terms of player skill and where these 11 players fill the team's formation) by around 13% on average, indicating more rest for key players without sacrificing performance. Additionally, MCTS achieved a 17% lower variance in expected points, leading to more stable season outcomes, and slightly better performance at the lower quartile.

Overall, MCTS offers a team selection strategy that parallels the greedy approach in rewards but significantly lowers player injuries, enhancing both competitive success and player welfare. This is particularly relevant given the financial impact of injuries on both teams and players. Comparison to the greedy approach underscores the tendency of managers to focus on short-term gains, often overlooking player welfare. Adopting our data-driven approach could thus improve long-term player health and challenge the prevalent short-term focus in soccer management.

### 7.4. **Experiment 4: Assessing Model Resilience in High-Risk Scenarios**

In this experiment, we evaluate the resilience of our MCTS model compared to the greedy strategy in scenarios with periods of inflated injury risk. This experiment is conducted to test the performance of the model in situations where the management of player welfare is critical such as intense cup runs, tournament soccer, or other high-risk sports like American football or rugby. We adjust the model to simulate heightened injury risks and limited player availability by multiplying the base probabilities by a constant. The outcomes of this are shown in Figure 7, where we measure performance improvement against the greedy strategy relative to the maximum possible expected points with an assumption of no injuries.

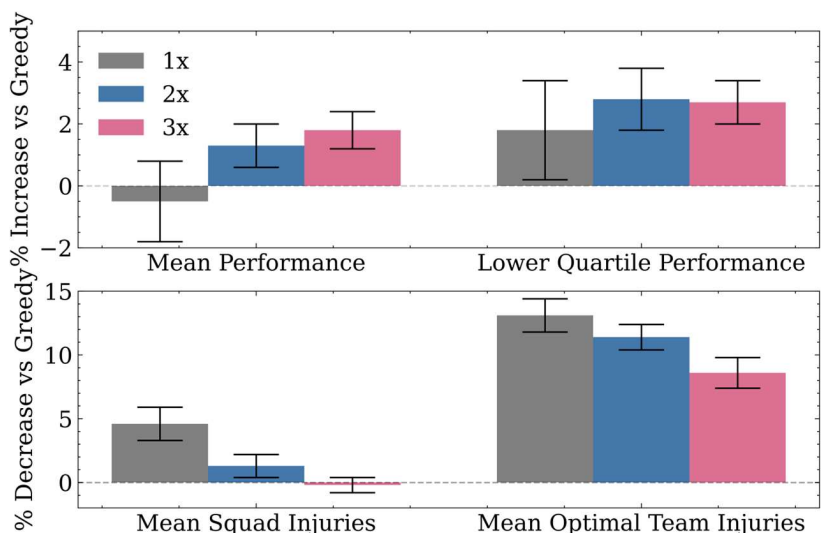

Figure 7 - MCTS results compared against the Greedy strategy with varying injury risk multipliers.

In high-risk scenarios, our MCTS model surpasses the greedy strategy in mean performance (i.e., season expected points), demonstrating its effectiveness in environments where managing and preventing injuries is crucial. We also find that the lower quartile of season performance is consistently improved against the Greedy strategy for all risk multipliers, highlighting the strengths of MCTS in reducing the risk of highly underachieving in a season. As the injury probability rises, the injury reduction margin compared to the greedy strategy narrows, possibly due to frequent injuries limiting proactive choices. Yet, MCTS consistently results in fewer injuries among the optimal team, maintaining its advantage in injury mitigation.

## 7.5. **Experiment 5: Team Financial Impact**

In this experiment, we focus on comparing the financial implications of injuries under the MCTS and Greedy strategies during the simulations of the 2018/19 EPL season. Building on the findings of Experiment 3, which established the parity in seasonal performance between the MCTS and Greedy strategies, we assume equivalent competition prize money and other financial rewards for both methods. Our primary metric for this experiment is the cost incurred due to player injuries, where money is spent inefficiently on player wages whilst they are injured and unavailable to play.

We sourced player wage data for the 2018/19 EPL season from Capology[6], allowing us to quantify the monetary impact of injuries. This is shown in Figure 8, highlighting the percentage reduction in wages spent on injured players when teams adopt the MCTS strategy as opposed to the Greedy approach. This comparison is presented for each club in the league, offering a comprehensive view of the financial benefits associated with the implementation of our MCTS strategy in terms of injury-related cost savings.

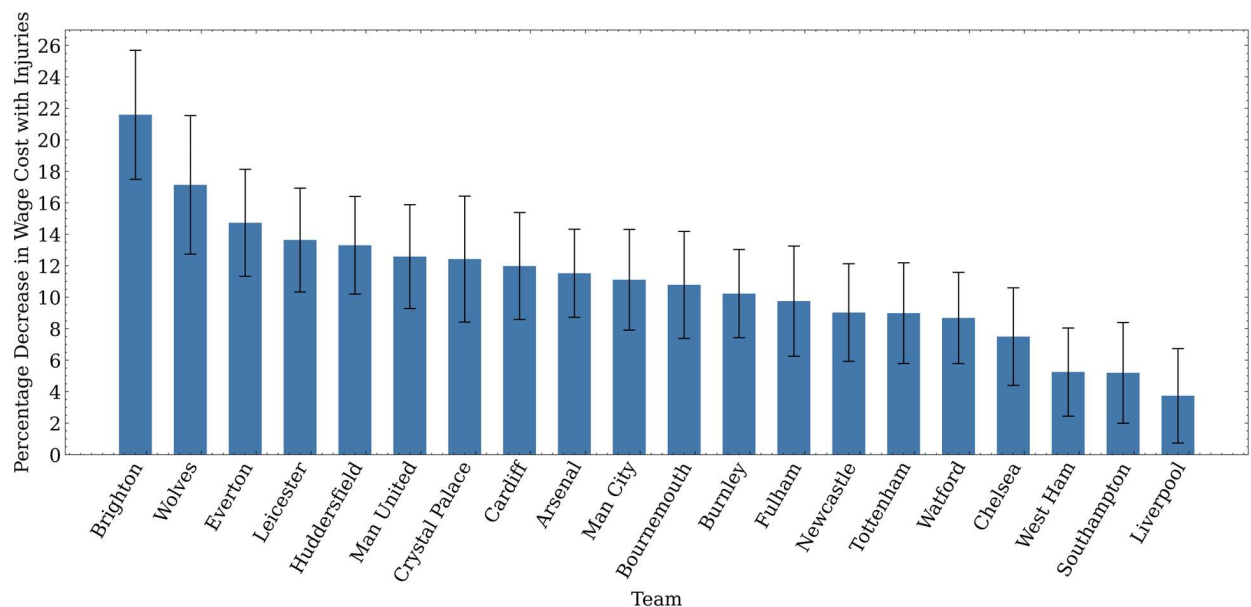

Figure 8 – The percentage reduction in wages spent on injured players for each team in the EPL over the 2018/19 season simulations using MCTS compared to the Greedy strategy.

Our analysis highlights that adopting the MCTS strategy for team selection would result in an average reduction of 11% in the money inefficiently spent on injured players across all clubs compared to Greedy. This translates to an average improvement of approximately £700,000 in player wages per club for the season. Notably, the clubs benefiting most from the MCTS strategy in terms of wage savings are Manchester United and Manchester City, where the reduction in wages inefficiently spent on injured players are projected as £1.88 million and £1.80 million respectively. These findings demonstrate that all clubs can achieve significantly improved efficiency in wage spending with regard to player injury by implementing the MCTS strategy, underlining its long-term financial advantages. We also highlight that these values are likely underexaggerated, as they do not consider the long-term effect of repeated injuries (as shown by the SHAP values in Figure 10) and the requirement for a larger, more expensive squad to deal with many player injuries.

---

[6] Wage data available at: https://www.capology.com/

## 7.6. **Experiment 6: Player Welfare Case Study**

In this experiment, we conducted a case study to assess the injury risk of a player throughout the 2018/19 EPL season. We chose N'Golo Kanté, a regular starter for Chelsea, for this study. The study compares Kanté's actual game participation in the 2018/19 season with a simulated season where team selections were made using our MCTS model. This comparison aims to demonstrate the benefits of long-term planning inherent in the MCTS strategy, particularly how increased rest periods can affect a player's injury risk over the course of a season. Figure 9 illustrates the variation in injury probability for Kanté across each gameweek under both the real-world selections and the MCTS-based selections.

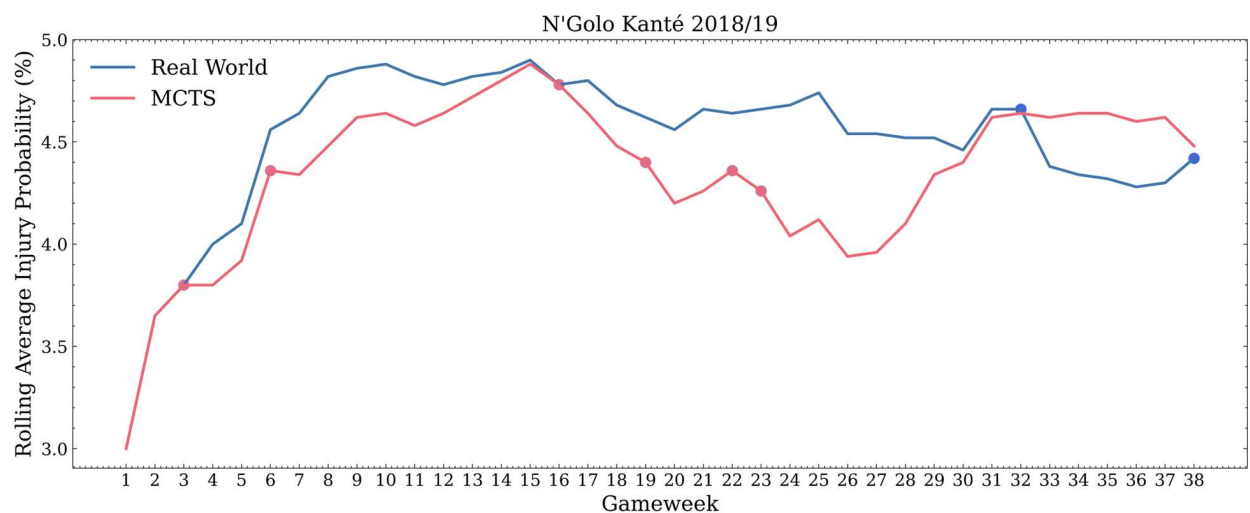

Figure 9 - Comparison between the rolling average injury probability (window size = 5) of N'Golo Kanté over the 2018/19 EPL season both using the matches played in the real world compared to the MCTS suggested actions in a season simulation. Dots represent rests and the player is selected for all gameweeks without dots.

Figure 9 illustrates the benefits of increased rest periods in reducing a player's injury risk throughout the season. The MCTS strategy effectively identifies optimal times for player rest, contributing to improved long-term player welfare. This is especially significant in light of Experiment 3's findings, which demonstrate that the MCTS strategy can maintain expected season points while enhancing player welfare.

The MCTS strategy shows a slight increase in injury risk towards the season's end, attributed to its focus on a single season where long-term injuries are less impactful as the season concludes. Extending the model to multiple seasons could further enhance its effectiveness in safeguarding long-term player welfare.

# 8. **Discussion**

In this paper, we have found that our MCTS model demonstrates greater stability in decision-making compared to the greedy strategy, indicated by its lower variance across simulations. This consistency is particularly beneficial in team sports for achieving specific objectives, such as avoiding relegation or qualifying for a continental competition. Interestingly, MCTS often reserves key players for matches where their impact is significant, rather than deploying them against much stronger teams with low win probabilities. This strategic use of players offers valuable insights for soccer team management.

Expanding our model to include strategies like workload distribution within a single game, especially through soccer substitutions, is a potential area for future research. This could involve planning multiple team selections within a single game, subject to substitution rules, to optimize performance in various scenarios, such as when leading comfortably.

The model presented in this paper, while initially applied to soccer due to data availability, has significant potential across various team sports. Its capacity to improve player welfare, decrease injuries, and optimize performance for diverse objectives, such as points per game or winning rates, underscores its wide-ranging applicability. This is particularly pertinent in sports with higher injury risks. Additional validation in various sports settings where data is accessible would further solidify the model's robustness.

The application of this model could be transformative in player welfare and injury management, offering coaches a data-driven way to balance performance with player health. This approach could not only reduce injuries and enhance team selections but also foster more sustainable team management practices through explainable injury risks.

Beyond reducing injuries, this approach promises significant financial benefits. By ensuring that highly paid players miss fewer games due to injuries, teams can optimize their financial investments. This can also potentially increase a player's earning potential by extending their career longevity. The model could therefore reshape team strategies and player development plans, ensuring that both the players' careers and the teams' financial health are managed more effectively over long time periods.

## 9. Conclusion

In conclusion, this paper presents a novel Monte-Carlo Tree Search (MCTS) algorithm for soccer team selection that prioritizes player health without compromising performance. Tested against a greedy strategy with 2018/19 English Premier League data, our MCTS approach demonstrated a significant reduction in injuries among key players by approximately 13%, while maintaining competitive season performance. This research offers a transformative perspective for the sports industry, showing that strategic, data-driven team selection can simultaneously protect players from injuries and achieve desired on-field results. It not only aids clubs in achieving long-term goals through enhanced player longevity but also promises financial benefits by lowering injury-related costs. Additionally, our approach, effective even in high-injury-risk environments, has potential applications in various team sports, suggesting a universal strategy for balancing competitive success and player welfare.

# References

[1]  O. Owoeye, M. VanderWay and I. Pike, "Reducing Injuries in Soccer (Football): an Umbrella Review of Best Evidence Across the Epidemiological Framework for Prevention," *Sports Medicine,* vol. 6, no. 46, 2020.

[2]  R. Beal, T. Norman and S. Ramchurn, "Artificial intelligence for team sports: a survey," *The knowledge engineering review,* vol. 34, 2019.

[3]  B. Hulin, T. Gabbett, D. Lawson, P. Caputi and J. Sampson, "The acute:chronic workload ratio predicts injury: high chronic workload may decrease injury risk in elite rugby league players," *British journal of sports medicine,* vol. 50, pp. 231-236, 2016.

[4]  K. Kucera, S. Marshall, D. Kirkendall and P. Marchak, "Injury history as a risk factor for incident injury in youth soccer," *British Journal of Sports Medicine,* vol. 39, pp. 462-466, 2005.

[5]  J. Orchard and J. Powell, "Risk of Knee and Ankle Sprains under Various Weather Conditions in American Football," *Medicine and Science in Sports and Exercise,* vol. 35, pp. 1118-23, 2003.

[6]  R. Beal, N. Changder, T. Norman and S. Ramchurn, "Learning the value of teamwork to form efficient teams," in *Proceedings of the AAAI Conference on Artificial Intelligence*, 2020.

[7]  T. Matthews, S. Ramchurn and G. Chalkiadakis, "Competing with Humans at Fantasy Football: Team Formation in Large Partially-Observable Domains," in *Proceedings of the AAAI Conference on Artificial Intelligence*, 2012.

[8]  G. Chalkiadakis and C. Boutillier, "Sequentially optimal repeated coalition formation under uncertainty," in *Autonomous Agents and Multi-Agent Systems*, 2012.

[9]  M. Gaston and M. desJardins, "Agent-organized networks for dynamic team formation," in *Proceedings of the fourth international joint conference on Autonomous agents and multiagent systems*, 2005.

[10]  T. Okimoto, N. Schwind, M. Clement, T. Ribeiro, K. Inoue and P. Marquis, "How to Form a Task-Oriented Robust Team," in *Proceedings of the 2015 International Conference on Autonomous Agents and Multiagent Systems*, 2015.

[11]  E. Demirovic, N. Schwind, T. Okimoto and K. Inoue, "Recoverable Team Formation: Building Teams Resilient to Change," in *Proceedings of the 17th International Conference on Autonomous Agents and Multi Agent Systems*, 2018.

[12]  T. Gunn and J. Anderson, "Dynamic heterogeneous team formation for robotic urban search and rescue," in *Journal of Computer and System Sciences*, 2015.

[13]  G. Everett, R. Beal, T. Matthews, J. Early, T. Norman and S. Ramchurn, "Inferring Player Location in Sports Matches: Multi-Agent Spatial Imputation from Limited Observations," in *Proceedings of the 2023 International Conference on Autonomous Agents and Multiagent Systems*, 2023.

[14] S. Lundberg and S.-I. Lee, "A unified approach to interpreting model predictions," in *Proceedings of the 31st International Conference on Neural Information Processing Systems*, 2017.

[15] M. Maher, "Modelling association football scores," *Statistica Neerlandica,* vol. 36, pp. 109-118, 1982.

[16] T. Decroos, L. Bransen, J. Van Haaren and J. Davis, "VAEP: An Objective Approach to Valuing On-the-Ball Actions in Soccer," in *Proceedings of the Twenty-Ninth International Joint Conference on Artificial Intelligence*, 2020.

[17] R. Beal, G. Chalkiadakis, T. Norman and S. Ramchurn, "Optimising Long-Term Outcomes using Real-World Fluent Objectives: An Application to Football," in *Proceedings of the 20th International Conference on Autonomous Agents and MultiAgent Systems*, 2021.

[18] D. Silver, A. Huang, C. Maddison, A. Guez, L. Sifre, G. van den Driessche, J. Schrittwieser, I. Antonoglou, V. Panneershelvam, M. Lanctot, S. Dieleman, D. Grewe, J. Nham, N. Kalchbrenner, I. Sutskever, T. Lillicrap, M. Leach, K. Kavukcuoglu, T. Graepel and D. Hassabis, "Mastering the game of Go with deep neural networks and tree search," *Nature,* vol. 529, p. 484–489, 2016.

[19] F. Wu and S. Ramchurn, "Monte-Carlo Tree Search for Scalable Coalition Formation," in *Proceedings of the 29th International Joint Conference on Artificial Intelligence*, 2020.

[20] P. Auer, N. Cesa-Bianchi and P. Fischer, "Finite-time Analysis of the Multiarmed Bandit Problem," *Machine Learning,* vol. 47, pp. 235-256, 2002.

[21] G. Chaslot, M. Winands, J. Van Den Herik and J. Uiterwijk, "Progressive Strategies for Monte-Carlo Tree Search," *New Mathematics and Natural Computation,* vol. 4, pp. 343-357, 2008.

[22] R. Coulom, "Computing Elo Ratings of Move Patterns in the Game of Go," *ICGA journal,* vol. 30, pp. 198-208, 2007.

# Appendix - Feature Importance Chart

Figure 10 shows all the features used in the player injury probability model and the contribution of each of these features towards player injury risks across our whole test dataset.

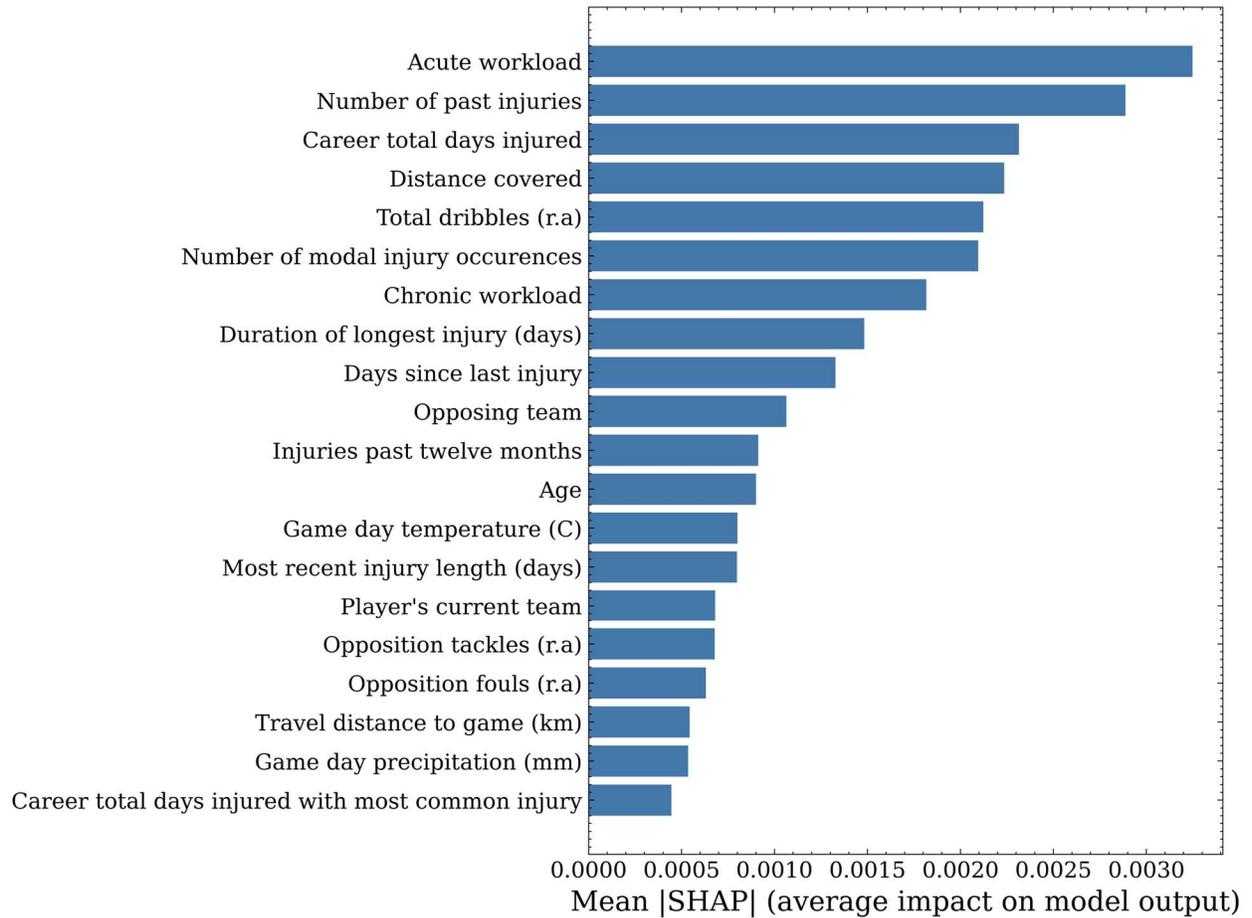

Figure 10 - Mean |SHAP| value for all player injury model predictions.